\documentclass[runningheads]{llncs}

\usepackage[year=2026]{eccv}

\usepackage{eccvabbrv}

\usepackage{graphicx}
\usepackage{booktabs}
\usepackage{multirow}
\usepackage{amsmath,amssymb,amsfonts}
\usepackage{enumitem}
\usepackage{subcaption}
\usepackage{algorithm}
\usepackage{algorithmic}
\usepackage{xcolor}

\usepackage[accsupp]{axessibility}  % Improves PDF readability for those with disabilities.

\usepackage[breaklinks,colorlinks,citecolor=eccvblue]{hyperref}

\begin{document}

% ---------------------------------------------------------------
\title{ExPLoRe: Expert Patch-Level Loss Routing for Multi-Objective Masked Image Modeling}

% TODO REVIEW: If the paper title is too long for the running head, you can set
% an abbreviated paper title here. If not, comment out.
\titlerunning{ExPLoRe: Expert Patch-Level Loss Routing for MIM}

\author{Konstantinos Georgiou \and
Maofeng Tang \and
Hairong Qi}

\authorrunning{K. Georgiou et al.}

\institute{Min H. Kao Department of Electrical Engineering and Computer Science,\\
The University of Tennessee, Knoxville, TN 37996, USA\\
\email{\{kgeorgio, mtang4\}@vols.utk.edu, hqi@utk.edu}}

\maketitle

\begin{abstract}
Multi-objective masked image modeling (MIM) combines complementary learning signals (token distillation, CLS alignment, and pixel reconstruction) but existing methods weight these objectives with global scalars, ignoring spatial heterogeneity across patches. We present ExPLoRe (Expert Patch-Level Loss Routing), which repurposes Soft Mixture of Experts (MoE) dispatch weights as learned, per-patch loss coefficients. The key mechanism is \emph{loss-coupling}: allowing loss gradients to flow through dispatch weights to the router enables content-dependent specialization, where different patches receive different emphases across objectives. A detach ablation confirms loss-coupling as the core mechanism, degrading performance by 1.6\% when gradients are blocked. On ImageNet-1K with ViT-Base, ExPLoRe improves over non-MoE baselines on two objective combinations (Token+CLS: +0.5\% k-NN, +4.4\% linear probe; Token+Pixel: +2.2\% k-NN), achieving 80.6\% linear probe and 85.3\% finetuning accuracy, competitive with published methods. For downstream transfer, we develop adaptation recipes (Freeze Routing, Expert Dropout, and Freeze Attention) that improve MoE finetuning by +1.5\% over the vanilla MoE, and close a 2.5--2.9 mIoU segmentation gap so that MoE models match or exceed non-MoE baselines on ADE20K.
\end{abstract}

\section{Introduction}
\label{sec:intro}

\begin{figure}[t]
\centering
\includegraphics[width=0.7\columnwidth]{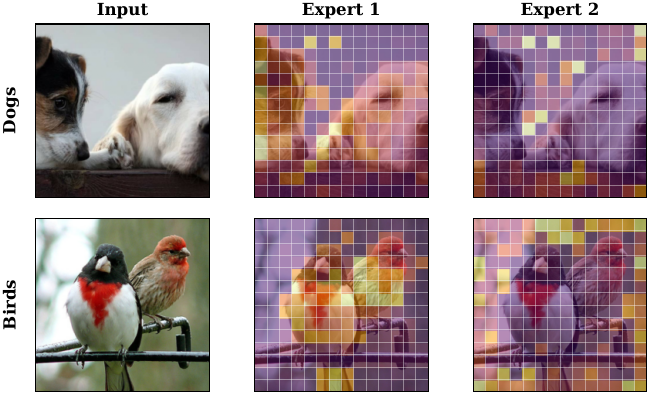}
\caption{\textbf{Expert dispatch-weight visualization.} Per-patch dispatch weights from a trained 2-expert ExPLoRe model overlaid on input images (warm\,=\,high weight, cool\,=\,low weight). The two experts learn complementary spatial specialization without explicit supervision: one expert assigns higher loss emphasis to foreground regions while the other focuses on background and context.}
\label{fig:expert_dispatch}
\end{figure}

Masked image modeling (MIM), inspired by BERT's masked language modeling~\cite{devlin2019bert}, has emerged as a powerful self-supervised learning paradigm for vision~\cite{bao2021beit, he2022masked}. Teacher-guided approaches~\cite{peng2022unified,hou2022milan,dong2023maskclip} leverage frozen feature extractors such as CLIP~\cite{radford2021learning} to provide semantic targets for student encoders. Recent works combine multiple learning signals including token-level distillation, CLS alignment, and pixel reconstruction; but how to weight these diverse objectives optimally remains an open question that is largely independent of the specific teacher used.

\textbf{The Patch-Level Heterogeneity Problem.}
Existing multi-task learning methods~\cite{chen2018gradnorm,lin2022rw} balance losses at the \emph{global} level, applying a single scalar weight per objective uniformly across all spatial locations. However, different image regions benefit from different learning signals: semantically rich patches are better served by teacher distillation, while texture-heavy background regions benefit from pixel reconstruction. Global methods cannot capture this spatial variation. As we demonstrate, gradient-based balancing (GradNorm~\cite{chen2018gradnorm}) can even fail catastrophically when objectives have spatially heterogeneous importance. To our knowledge, no prior work addresses \emph{patch-level} loss weighting for masked image modeling.

%\textbf{Our Solution: ExPLoRe.}
We present ExPLoRe\footnote{Code available at \url{https://github.com/aicip/ExPLoRe}.} (\textbf{Ex}pert \textbf{P}atch-\textbf{L}evel L\textbf{o}ss \textbf{R}outing), which addresses patch-level heterogeneity by repurposing Soft Mixture of Experts (Soft-MoE)~\cite{softmoe} dispatch weights as learned, per-patch loss coefficients. Unlike traditional MoE methods that use routing for capacity scaling~\cite{shazeer2017outrageously,fedus2021switch,vmoe} or task-specific expert assignment~\cite{chen2023adamvmoe,yang2025astrea}, ExPLoRe uses continuous, sample-adaptive weighting through a \emph{loss-coupling} mechanism: gradients from each loss objective flow back through the dispatch weights to the router, enabling the model to learn content-dependent specialization. We evaluate two objective combinations (\emph{Token+CLS} and \emph{Token+Pixel}) with expert scaling from 2 to 64 under sparse encoding~\cite{he2022masked}, and develop downstream transfer recipes for competitive finetuning. Figure~\ref{fig:expert_dispatch} visualizes the resulting routing patterns. 

ExPLoRe provides content-dependent per-patch loss emphasis that prior methods lack; competitive results across classification and dense prediction demonstrate this mechanism does not sacrifice representation quality.
 Our contributions center on a single mechanism with two supporting analyses:
\begin{itemize}[itemsep=2pt,topsep=3pt]
    \item \textbf{Core contribution: dispatch-weight loss weighting.} We introduce a mechanism that repurposes Soft-MoE dispatch weights as per-patch loss coefficients for multi-objective MIM. Loss-coupling, where loss gradients flow through dispatch weights to update the router, is the core innovation enabling learned per-patch specialization. We demonstrate consistent improvements over non-MoE baselines across two objective combinations, achieving 80.6\% linear probe accuracy competitive with published methods (Table~\ref{tab:comparison}).

    \item \textbf{Routing dynamics analysis.} We empirically characterize routing behavior: entropy regularization governs a sharp transition between stable routing and catastrophic collapse; optimal regularization is expert-count dependent; %($\lambda\!=\!0.5$ helps 2 experts, hurts 64); 
    and a detach ablation validates loss-coupling as the mechanism driving specialization (Sections~\ref{sec:ablations} and \ref{sec:routing_analysis}).

    \item \textbf{Downstream transfer recipes.} We develop complementary adaptation strategies, Freeze Routing (FR), Expert Dropout (ExD), and Freeze Attention (FA), that compose naturally. The combined FR+FA+ExD recipe exceeds the non-MoE baseline on both classification (+0.5\%) and semantic segmentation, closing a 2.5--2.9 mIoU gap (Section~\ref{sec:semseg_results}).
\end{itemize}

\section{Related Work}
\label{sec:related}

\textbf{Masked Image Modeling.}
MIM extends masked language modeling (MLM) from NLP to vision. BEiT~\cite{bao2021beit} uses \emph{dense encoders} processing all patches with mask tokens, while MAE~\cite{he2022masked} uses \emph{sparse encoders} processing only visible patches with shuffling for efficiency. SimMIM~\cite{xie2022simmim} uses simple linear projections, iBOT~\cite{zhou2021ibot} combines masking with self-distillation, and MaskFeat~\cite{wei2022masked} uses HOG features as targets.

Self-distillation approaches include data2vec~\cite{baevski2022data2vec,baevski2023data2vec2} (EMA representations), CAE/CAEv2~\cite{chen2024context,zhang2024caev2} (context autoencoding), SdAE~\cite{chen2022sdae} (pixel + self-distillation), and BootMAE~\cite{dong2022bootstrapped} (bootstrapping).

\textbf{Teacher-Guided MIM with CLIP.}
CLIP~\cite{radford2021learning} enables MIM methods with semantic targets. MaskDistill~\cite{peng2022unified} provides a unified view of masked image modeling, formulating MIM as $\mathcal{L}(\mathcal{N}(\mathcal{T}(I)), \mathcal{H}(\mathcal{S}(I_m)))$ with teacher $\mathcal{T}$, student $\mathcal{S}$, projector heads $\mathcal{N}$/$\mathcal{H}$, and shows CLIP features outperform pixels or discrete tokens as reconstruction targets. Our base architecture builds upon MaskDistill's framework. BEiT v2~\cite{peng2022beit} combines tokenizers with CLIP distillation. MILAN~\cite{hou2022milan} uses CLIP attention for semantic masking and combines CLS with patch distillation. MaskCLIP~\cite{dong2023maskclip} explores masked self-distillation within CLIP.

\textbf{Multi-Task Learning and Loss Balancing.}
Multiple methods balance competing objectives: uncertainty weighting~\cite{kendall2018multi} models task-dependent uncertainty, GradNorm~\cite{chen2018gradnorm} normalizes gradient magnitudes, PCGrad~\cite{yu2020gradient} projects conflicting gradients, MGDA~\cite{sener2018multi} seeks Pareto-optimal solutions, and random weighting~\cite{lin2022rw} samples task weights stochastically. Crucially, all operate at the \emph{global} level, applying a single scalar weight per objective uniformly across all spatial locations. This ignores the spatial structure of images, where different regions may benefit from different loss emphasis. To our knowledge, no prior work addresses per-patch adaptive loss weighting for masked image modeling, which we identify as a key gap.

\textbf{Mixture of Experts.}
Sparse MoE methods~\cite{shazeer2017outrageously,fedus2021switch} route tokens to expert subsets for conditional computation but introduce load balancing challenges. Soft-MoE~\cite{softmoe} addresses this through fully-differentiable soft routing, where dispatch and combine weights computed via softmax avoid discrete decisions and their associated training difficulties. V-MoE~\cite{vmoe} applied sparse MoE to vision transformers; subsequent work~\cite{liu2024routers,han2024vimoe} compared routing strategies, and multi-task applications~\cite{chen2023adamvmoe,yang2025astrea} demonstrated MoE effectiveness in vision. CR-MoE~\cite{jiang2024crmoe} first applied MoE to self-supervised learning (contrastive), discovering routing consistency challenges. However, applications to masked image modeling with dynamic task weighting have not been investigated. MoE downstream transfer remains understudied; ViMoE~\cite{han2024vimoe} finds that only 2--5 MoE layers suffice but focuses on supervised training. In contrast to prior work on capacity scaling~\cite{shazeer2017outrageously,fedus2021switch,vmoe} or task-specific expert assignment~\cite{chen2023adamvmoe,yang2025astrea}, ExPLoRe repurposes Soft-MoE dispatch weights for \emph{per-patch loss weighting} through continuous, loss-coupled routing, and addresses the MoE transfer gap with dedicated adaptation recipes (Section~\ref{sec:recipe_results}).

\section{Method}
\label{sec:method}

\begin{figure}[!htbp]
\centering
\includegraphics[width=0.75\textwidth]{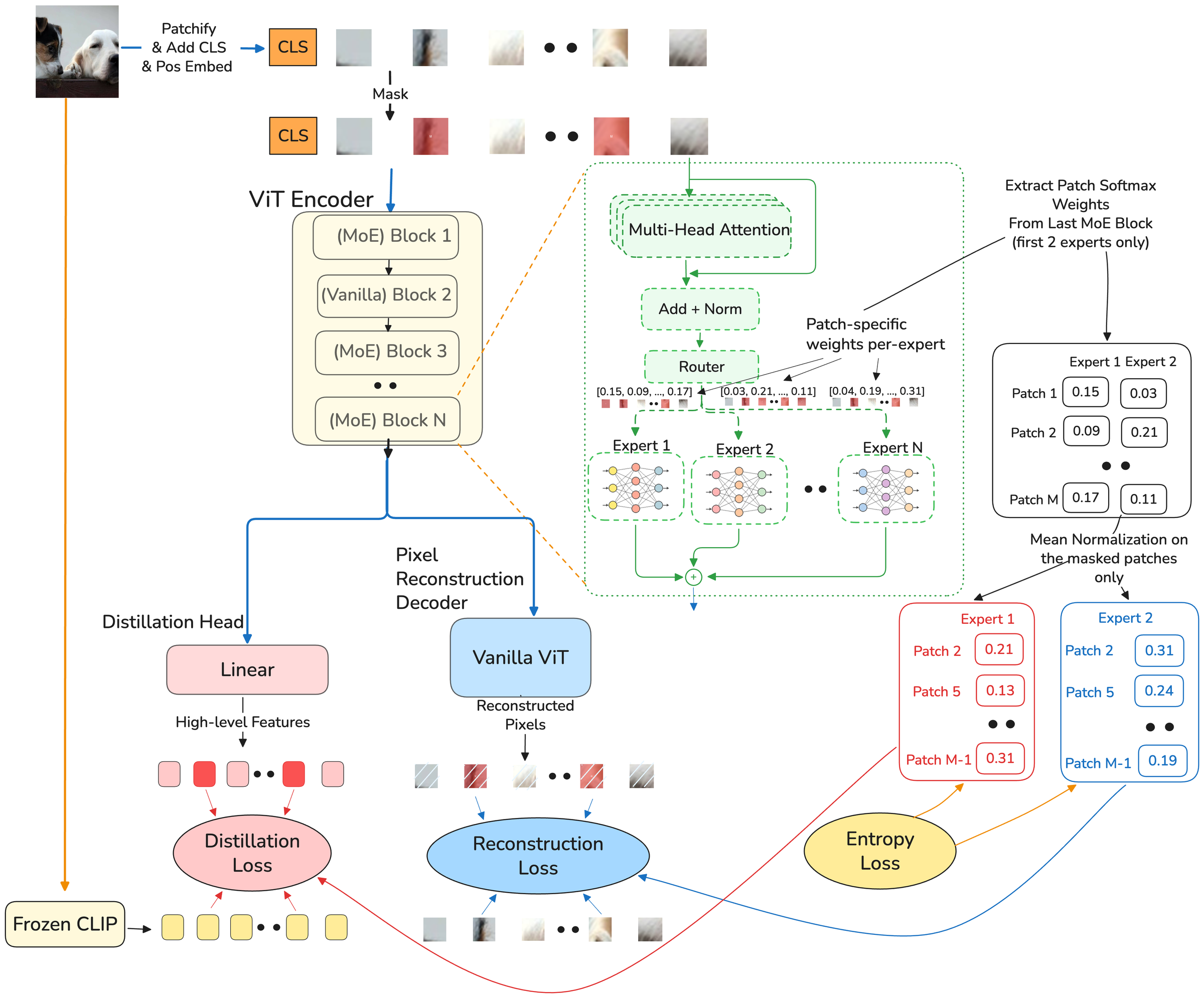}
\caption{\textbf{ExPLoRe Framework Overview.} Soft Mixture of Experts (Soft-MoE) is integrated into the student encoder for patch-level adaptive loss weighting. The student encoder (ViT-Base with alternating MoE blocks at layers \{1,3,5,7,9,11\}) processes patches while a frozen CLIP teacher provides semantic targets. Soft-MoE dispatch weights $\mathbf{D}$ serve as per-patch loss coefficients: each expert weights a different training objective. The two routing weight types play distinct roles: dispatch weights $\mathbf{D}$ (normalized over patches per expert) form the loss-coupling pathway that carries loss gradients back to the router, whereas combine weights $\mathbf{C}$ (normalized over experts per patch) only mix expert outputs in the forward pass. Loss-coupling, where loss gradients flow through $\mathbf{D}$ to the router, is the key mechanism enabling learned specialization.}
\label{fig:model}
\end{figure}

ExPLoRe uses Soft-MoE to learn per-patch loss weighting in multi-objective masked image modeling, repurposing dispatch weights as dynamic loss coefficients (Figure~\ref{fig:model}). Given an input image divided into patches with a binary mask partitioning them into visible set $\mathcal{V}$ and masked set $\mathcal{M}$, the framework consists of: (1)~a student encoder processing visible patches, (2)~a frozen CLIP teacher providing semantic targets, and (3)~optional auxiliary networks (decoder for pixel reconstruction, projection heads for CLS alignment).

\subsection{Multi-Objective Learning Framework}

Our framework combines three complementary objectives at different spatial scales. Per-image losses are defined below; MoE-weighted variants (Section~\ref{sec:moe_integration}) average over the batch.

\textbf{Token-Level Distillation}: Distills CLIP teacher features $\mathbf{t}$ into student predictions $\hat{\mathbf{t}}$ using Huber loss ($\beta\!=\!1.0$) with layer-normalized teacher features:
\begin{equation}
\label{eq:loss_token}
\mathcal{L}_\text{token} = \frac{1}{|\mathcal{M}|} \sum_{i \in \mathcal{M}} \ell_\text{huber}(\hat{\mathbf{t}}_i, \text{LN}(\mathbf{t}_i))
\end{equation}
\textbf{Global CLS Alignment}: Aligns CLS tokens of student and teacher:
\begin{equation}
\mathcal{L}_\text{cls} = 1 - \frac{\hat{\mathbf{z}}_\text{cls} \cdot \mathbf{t}_\text{cls}}{\|\hat{\mathbf{z}}_\text{cls}\|\,\|\mathbf{t}_\text{cls}\|}
\end{equation}

\textbf{Pixel Reconstruction}: A decoder reconstructs normalized pixel values:
\begin{equation}
\label{eq:loss_pixel}
\mathcal{L}_\text{pixel} = \frac{1}{|\mathcal{M}|} \sum_{i \in \mathcal{M}} \|\hat{\mathbf{x}}_i - \bar{\mathbf{x}}_i\|_2^2
\end{equation}
where $\hat{\mathbf{x}} = d_\omega(\mathbf{z})$ are reconstructed patches and $\bar{\mathbf{x}}$ are normalized targets. ExPLoRe uses Soft MoE dispatch weights as learned per-patch loss coefficients for the combined objective (Section~\ref{sec:moe_integration}).

\textbf{Encoding Strategy and Decoder.}
\label{sec:arch_variations}
We adopt sparse encoding (MAE-style~\cite{he2022masked}) for our main model, processing only visible patches for efficiency. Sparse encoding yields dispatch weights only for visible patches, determining which losses can be MoE-weighted (Section~\ref{sec:moe_integration}). Dense encoding (BEiT-style~\cite{bao2021beit}) is evaluated as an ablation. A lightweight 8-block decoder reconstructs masked patches; full details in supplementary Section~E.

\subsection{Soft Mixture of Experts for Patch-Level Dynamic Loss Weighting}

Global methods apply uniform weights across patches, but different patches benefit from different signals: semantic regions need distillation, texture regions need reconstruction. We integrate Soft-MoE~\cite{softmoe} into the student encoder for learned patch-level adaptive weighting (Figure~\ref{fig:model}). The central idea is that Soft-MoE produces two types of weights from shared logits: \emph{dispatch weights} $\mathbf{D}$ (softmax over patches, normalized per expert) that determine how patches contribute to each expert, and \emph{combine weights} $\mathbf{C}$ (softmax over experts, normalized per patch) that determine the output mixture. We repurpose dispatch weights, rather than combine weights, as loss coefficients, for reasons detailed below.

\textbf{Architecture and Routing Mechanism}: We adopt the Soft-MoE formulation~\cite{softmoe} (full derivation in supplementary Section~G). Soft-MoE replaces selected MLP layers with $E$ expert networks, each a standard MLP with independent parameters. Given patch representations $\mathbf{X} \in \mathbb{R}^{B \times N \times d}$, learnable expert parameters $\mathbf{\Phi} \in \mathbb{R}^{d \times E}$, and a learned scale $s$, routing logits are computed as:
\begin{equation}
\mathbf{L} = s \cdot \frac{\mathbf{X}}{\|\mathbf{X}\|_2} \cdot \frac{\mathbf{\Phi}}{\|\mathbf{\Phi}\|_2}
\end{equation}
These shared logits yield two weight types via softmax along different dimensions: \emph{dispatch weights} $\mathbf{D} = \text{softmax}_\text{patches}(\mathbf{L})$ with $\sum_{n} \mathbf{D}_{b,n,e} = 1$ per expert, and \emph{combine weights} $\mathbf{C} = \text{softmax}_\text{experts}(\mathbf{L})$ with $\sum_{e} \mathbf{C}_{b,n,e} = 1$ per patch. Dispatch weights aggregate patches into expert inputs $\mathbf{S}_\text{in} = \mathbf{D}^\top \mathbf{X}$, each expert processes its slot, and combine weights reconstruct per-patch outputs $\mathbf{Y} = \mathbf{C} \cdot \mathbf{S}_\text{out}$. The critical distinction for loss weighting is that dispatch weights are normalized over patches (per expert) while combine weights are normalized over experts (per patch).

\textbf{Dispatch-Weight-Based Loss Weighting}: Our key innovation is using dispatch weights $\mathbf{D}$ (not combine weights) for patch-level loss weighting. Unlike combine weights which represent per-token expert mixtures with constraint $\sum_e \mathbf{C}_{b,n,e} = 1$ (normalized over experts for each patch), dispatch weights represent how much each patch contributes to each expert with constraint $\sum_n \mathbf{D}_{b,n,e} = 1$ (normalized over patches for each expert). This expert-centric perspective prevents a critical degeneracy issue: with combine weights, the router could minimize loss by setting all weights to zero for difficult patches, causing the loss to vanish. With dispatch weights, the normalization constraint over patches prevents this collapse, made concrete by the per-image normalization in Eq.~\ref{eq:moe_token} below. The router must distribute each expert's "attention" across patches and cannot zero out the entire loss.

For our main configuration (Token+CLS, sparse encoding), Expert~0 weights the token distillation loss on visible patches $\mathcal{V}$. Each patch's contribution is scaled by its dispatch weight, averaged over the batch:
\begin{equation}
\label{eq:moe_token}
\mathcal{L}_\text{token}^\text{MoE} = \frac{1}{B} \sum_{b=1}^B \frac{\sum_{i \in \mathcal{V}} \mathbf{D}_{b,i,0} \cdot \ell_\text{huber}(\hat{\mathbf{t}}_{b,i}, \text{LN}(\mathbf{t}_{b,i}))}{\sum_{i \in \mathcal{V}} \mathbf{D}_{b,i,0}}
\end{equation}

Per-image normalization (dividing by the sum of dispatch weights) computes a weighted average for each image independently, preventing the router from minimizing loss by scaling weights down across the batch.

When $E > K$ (more experts than loss-weighted objectives), only $K$ experts receive direct loss-coupling; the rest provide MoE capacity through forward-pass gradients and entropy regularization (supplementary Section~H). The pixel loss variant applies analogous weighting (supplementary Section~D); this requires dense encoding since sparse mode produces dispatch weights only for visible patches $\mathcal{V}$, while pixel loss is computed on masked patches $\mathcal{M}$.

\textbf{Entropy Regularization for Uniform Token Distribution}: While dispatch weights prevent loss degeneracy, they don't guarantee balanced expert utilization. Without regularization, one expert might receive contributions from all patches while another receives none, leading to expert collapse. Per-expert entropy regularization encourages uniform token distributions.

For each expert $e$, we compute the entropy of its dispatch-weight distribution across patches:
\begin{equation}
H_e = -\frac{1}{B}\sum_{b=1}^B \sum_{n=1}^N p_{b,n,e} \log p_{b,n,e}
\end{equation}
where $p_{b,n,e} = \mathbf{D}_{b,n,e}$, which already forms a distribution over patches per expert since Soft-MoE normalizes dispatch weights over the token axis ($\sum_n \mathbf{D}_{b,n,e} = 1$). Higher entropy indicates a more uniform distribution across patches.

The entropy loss encourages high entropy (uniformity) through minimization:
\begin{equation}
\label{eq:entropy}
\mathcal{L}_\text{entropy} = -\sum_{e=0}^{E-1} \lambda_e \cdot H_e
\end{equation}
where $\lambda_e$ are per-expert weights. For Token+CLS, we use a symmetric weight $\lambda$ for all experts; through hyperparameter search, we find $\lambda\!=\!5.0$ provides stable training as default, while $\lambda\!=\!0.5$ is optimal for 2 experts. Critically, optimal entropy weight is \emph{expert-count dependent}: $\lambda\!=\!0.5$ improves 2-expert routing but degrades 64-expert performance (Section~\ref{sec:routing_analysis}). Token+Pixel requires asymmetric per-expert weights (supplementary Section~C). Regularization is computed at the last MoE block (Block~11), validated via block selection ablation (Section~\ref{sec:ablations}). Entropy regularization and loss-coupling are opposing forces: entropy pushes dispatch weights toward uniformity (no specialization), while loss-coupling pushes them toward content-dependent specialization; the coefficient $\lambda$ governs the trade-off, with collapse at $\lambda\!=\!0.01$, stable specialization at the default, and over-regularization (degradation) at $\lambda\!\gg\!5$ (Section~\ref{sec:routing_analysis}).

\textbf{MoE Placement and Implementation}:
\label{sec:moe_integration}
MoE replaces MLP layers in alternating blocks \{1, 3, 5, 7, 9, 11\} of ViT-Base, balancing capacity (6 MoE blocks) with efficiency (6 standard blocks) at approximately 1.5$\times$ computation. Expert networks use hidden dimension 3072 with GELU activation; $E\!=\!2$ experts increase parameters by 35\% (86M$\to$116M), while $E\!=\!64$ reaches 1.86B. We use $S\!=\!1$ slot per expert for direct expert-to-loss mapping. Full design rationale, memory analysis, and dense/sparse integration details are in the supplementary material (Sections~D--G).

\textbf{Loss-Coupling Mechanism}: Because dispatch weights are differentiable, gradients from each loss objective flow back through $\mathbf{D}$ to the router parameters $\mathbf{\Phi}$ and $s$, creating a \emph{loss-coupling} effect that enables learned specialization. A detach ablation confirms this as the core mechanism (Section~\ref{sec:ablations}).

\textbf{Total Training Objective}: Consolidating the per-objective losses (Eqs.~\ref{eq:loss_token}--\ref{eq:loss_pixel}) under dispatch-weight coupling, the full training objective is
\begin{equation}
\label{eq:total}
\mathcal{L}_\text{total} = \sum_{e} w_e \cdot \frac{1}{|\mathcal{P}_e|} \sum_{n \in \mathcal{P}_e} \mathbf{D}_{b^\star\!, n, e} \cdot \mathcal{L}_e(n),
\end{equation}
where $\mathcal{L}_e(n)$ is the per-patch loss for objective $e$ over its patch set $\mathcal{P}_e$, $w_e$ the global per-objective weight, and $\mathbf{D}_{b^\star\!, n, e}$ the dispatch weight from loss block $b^\star$ (Eq.~\ref{eq:moe_token}; $\mathbf{D}\!\equiv\!1$ for un-coupled objectives). The product $\mathbf{D}_{b^\star\!, n, e}\,\mathcal{L}_e(n)$ is the per-patch loss coefficient named in the abstract; entropy regularization (Eq.~\ref{eq:entropy}) is added during training.

\section{Experiments}
\label{sec:experiments}

\subsection{Experimental Setup}

\textbf{Dataset.} All experiments use ImageNet-1K~\cite{deng2009imagenet} (1.28M training images, 1000 classes).

\textbf{Pretraining.} ViT-Base/16 student encoder with frozen CLIP ViT-B/16 teacher, 300 epochs, batch size 4096 on 4$\times$ H100 GPUs with block masking at 40\% ratio and sparse encoding. Full hyperparameters are in supplementary Tables~A1--A2.

\textbf{Evaluation.} We evaluate across four complementary tasks: (1)~\textbf{k-NN} ($k\!=\!20$, cosine similarity) as a training-free representation quality metric; (2)~\textbf{Linear probe} on frozen [CLS] features; (3)~\textbf{ImageNet finetuning} with layer-wise LR decay~\cite{bao2021beit}; and (4)~\textbf{Semantic segmentation} on ADE20K~\cite{zhou2017scene} with UperNet~\cite{xiao2018unified}. MoE models use the adaptation recipes from Section~\ref{sec:recipe_results}. Full evaluation protocols are in supplementary Section~A.3.

\textbf{Model Configurations.} We evaluate two multi-objective combinations, summarized in Table~\ref{tab:config_summary}. Token+CLS is our primary configuration because it achieves the highest downstream accuracy and supports the full expert scaling range (2--64) without the inverted scaling observed in Token+Pixel at 64 experts. %ExPLoRe is applied to both, demonstrating generality across objective combinations.

\begin{table}[t]
\centering
\small
\caption{\textbf{MoE loss weighting configurations.} Which losses receive dispatch-weight modulation in each configuration. In sparse encoding, dispatch weights exist only for visible patches, so pixel loss (computed on masked patches) cannot be MoE-weighted.}
\label{tab:config_summary}
\begin{tabular}{lcccc}
\toprule
Configuration & Encoding & Token Loss & CLS Loss & Pixel Loss \\
\midrule
Token+CLS & Sparse & MoE (Exp.\ 0) & Global $w_\text{cls}$ & --- \\
Token+Pixel & Sparse & MoE (Exp.\ 0) & --- & Uniform \\
\bottomrule
\end{tabular}
\end{table}

%% The Multi-Objective Challenge (merged into Setup)
\label{sec:multi_obj_challenge}

\textbf{The Multi-Objective Challenge.} Existing global weighting methods fail on spatially heterogeneous MIM objectives. Despite an extensive hyperparameter search (supplementary Table~A3), GradNorm~\cite{chen2018gradnorm} cannot match its own static baseline (72.8\% vs.\ 74.2\% k-NN for Token+CLS) and collapses entirely on Token+Pixel (32.5\%), underperforming even its own dense baseline. Random loss weighting (RLW)~\cite{lin2022rw} helps modestly (75.8\% k-NN) but applies uniform per-patch weights, unable to capture spatial variation.

These failures motivate per-patch loss weighting: different image regions benefit from different loss emphasis, and global methods that apply a single scalar weight per objective cannot capture this spatial variation.

%% ─────────────────────────────────────────────────────────────
%% §4.3  MoE Mechanism Validation
%% Story: Our approach works → dispatch prevents collapse, loss-coupling is the key
%% ─────────────────────────────────────────────────────────────
\subsection{MoE Mechanism Validation}
\label{sec:ablations}
\label{sec:routing_analysis}

Having established the need for per-patch loss weighting, we validate the core design decisions of our MoE approach. Table~\ref{tab:ablation} summarizes ablations on the 2-expert Token+CLS configuration.

\textbf{Dispatch weights prevent degeneracy.} Combine-weight loss weighting collapses to 2.1\% k-NN ($-$73.3 points), confirming the dispatch-weight design choice (Section~\ref{sec:moe_integration}).

\textbf{Loss-coupling is the core mechanism.} We validate loss-coupling through a detach ablation: computing MoE-weighted losses with dispatch weights detached from the computation graph (i.e., treated as fixed constants during backpropagation so that loss gradients cannot update the router). This degrades k-NN by 1.6\% (Table~\ref{tab:ablation}), confirming that the router's ability to receive gradient signal from loss objectives is essential for learned specialization. The result is reinforced by the 2-expert unweighted variant (74.1\% k-NN, Table~\ref{tab:scaling}), which \emph{underperforms} the non-MoE baseline (75.7\%) by 1.6\%: MoE capacity alone is not merely insufficient but actively harmful without loss-coupling, making the coupling mechanism essential rather than incremental.

\textbf{Importance loss targets the wrong distribution.} Importance-based regularization ($w\!=\!2$) penalizes dispatch-weight imbalance across experts, but Soft-MoE's softmax already ensures balanced dispatch. The resulting scale parameter collapse ($s\!: 0.90 \to 0.22$ at the loss block) degrades performance by 0.1\%. Supplementary Section~D provides visualizations of dispatch-weight patterns across all six MoE blocks, confirming spatially coherent, content-dependent routing.

% \begin{table}[t]
% \centering
% \small
% \caption{\textbf{MoE design ablations} on 2-expert Token+CLS (k-NN@20). $\Delta$ relative to baseline ($\lambda\!=\!5.0$, Block~11). Top: entropy regularization sweep. Middle: mechanism ablations. Bottom: loss block selection.}
% \label{tab:ablation}
% \begin{tabular}{lcc}
% \toprule
% Configuration & k-NN@20 & $\Delta$ \\
% \midrule
% \multicolumn{3}{c}{\textit{Entropy Regularization}} \\
% \midrule
% $\lambda = 0.5$ (optimal 2-exp) & \textbf{75.5} & +0.1 \\
% $\lambda = 5.0$ (default) & 75.4 & --- \\
% $\lambda = 0.1$ & 75.3 & $-$0.1 \\
% $\lambda = 0.01$ (collapsed) & 66.9 & $-$8.5 \\
% \midrule
% \multicolumn{3}{c}{\textit{Mechanism Ablations}} \\
% \midrule
% Combine weight reg.\ ($\lambda\!=\!0.5$) & 75.4 & +0.0 \\
% Importance loss ($w\!=\!2$) & 75.3 & $-$0.1 \\
% Detach loss weights & 73.8 & $-$1.6 \\
% Combine weights (no entropy) & 2.1 & $-$73.3 \\
% \midrule
% \multicolumn{3}{c}{\textit{Loss Block Selection}} \\
% \midrule
% Block 11 (default) & 75.4 & --- \\
% Block 9 & 75.3 & $-$0.1 \\
% Block 7 & 75.4 & +0.0 \\
% Block 5 & 75.2 & $-$0.2 \\
% \bottomrule
% \end{tabular}
% \end{table}

\begin{table}[t]
\centering
\small
\caption{\textbf{MoE design ablations} on 2-expert Token+CLS (k-NN@20). $\Delta$ relative to baseline ($\lambda\!=\!5.0$, Block~11).}
\label{tab:ablation}
\setlength{\tabcolsep}{1.5pt}
\begin{tabular}{lcccc}
\toprule
\textbf{Entropy Reg.} &
$\lambda\!=\!0.5$ & $\lambda\!=\!5.0$ & $\lambda\!=\!0.1$ & $\lambda\!=\!0.01$ \\
\midrule
k-NN@20 & \textbf{75.5} & 75.4 & 75.3 & 66.9 \\
$\Delta$ & +0.1 & --- & $-$0.1 & $-$8.5 \\
\midrule
\textbf{Mech. Ablations} &
Combine reg. & Importance ($w\!=\!2$). & Detach weights. & No entropy \\
\midrule
k-NN@20 & 75.4 & 75.3 & 73.8 & 2.1 \\
$\Delta$ & +0.0 & $-$0.1 & $-$1.6 & $-$73.3 \\
\midrule
\textbf{Loss Block Sel.} &
Block 11 & Block 9 & Block 7 & Block 5 \\
\midrule
k-NN@20 & 75.4 & 75.3 & 75.4 & 75.2 \\
$\Delta$ & --- & $-$0.1 & +0.0 & $-$0.2 \\
\bottomrule
\end{tabular}
\end{table}

%% ─────────────────────────────────────────────────────────────
%% §4.4  Expert Scaling Analysis
%% Story: The mechanism scales smoothly from 2 to 64 experts
%% ─────────────────────────────────────────────────────────────
\subsection{Expert Scaling Analysis}
\label{sec:scaling}

\begin{figure}[t]
\centering
\includegraphics[width=0.75\columnwidth]{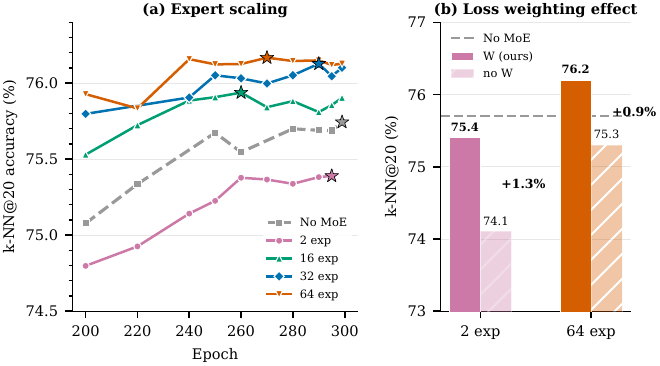}
\caption{\textbf{Expert scaling and mechanism isolation} (Token+CLS). \textbf{(a)}~k-NN@20 trajectories over epochs 200--300 for No~MoE and 2/16/32/64-expert configurations (all with dispatch-weight loss weighting). Stars mark peak accuracy per configuration; more experts yield higher final accuracy. \textbf{(b)}~Mechanism isolation: weighted (W) vs.\ unweighted (no~W) at 2 and 64 experts. Dispatch-weight loss weighting contributes +1.3\% at 2 experts and +0.9\% at 64 experts at matched parameter counts, confirming the mechanism's effect beyond the parameter contribution.}
\label{fig:expert_scaling}
\end{figure}

Table~\ref{tab:scaling} and Figure~\ref{fig:expert_scaling} present pretraining results across expert counts for both objective combinations. We include our MaskDistill~\cite{peng2022unified} reproduction (token distillation only, dense encoding) as a reference point. For \emph{Token+CLS}, starting from this MaskDistill baseline (75.6\% k-NN), adding CLS alignment without MoE improves k-NN to 75.7\%. Introducing 2-expert ExPLoRe with dispatch-weight loss weighting reveals an instructive pattern: k-NN decreases slightly to 75.4\%, but linear probe accuracy jumps to \textbf{79.6\%} (+3.4\% over No MoE), indicating that the routing mechanism reshapes the feature space toward improved linear separability.

k-NN scales monotonically from 2 to 64 experts, reaching \textbf{76.2\%} at 64 experts, with smooth convergence across all configurations (Figure~\ref{fig:expert_scaling}a). Linear probe, evaluated at the 2- and 64-expert endpoints, shows the same trend: 79.6\% $\to$ \textbf{80.6\%}. Critically, comparing weighted vs.\ unweighted variants \emph{at the same expert count} isolates loss weighting from the parameter effect (Figure~\ref{fig:expert_scaling}b). The 2-expert unweighted variant (74.1\%) demonstrates that loss weighting provides the +1.3\% gain that makes ExPLoRe competitive. The 64-expert unweighted variant (75.3\%) confirms that loss weighting contributes +0.9\% at this scale. The relative gain decreases from +1.3\% (2 experts) to +0.9\% (64 experts) as inherent diversity from more expert parameters provides partial dispatch variation independently of loss-coupling.

For \emph{Token+Pixel}: no MoE (71.5\%) $\to$ 32-expert without dispatch weighting (73.0\%, +1.5\%) $\to$ 32-expert with dispatch weighting (\textbf{73.7\%}, +2.2\% total). Dispatch weighting adds +0.7\% over the unweighted variant at 32 experts, confirming the mechanism's effectiveness across objective combinations. This +2.2\% total improvement is substantially larger than Token+CLS's +0.5\%, indicating that dispatch-weight loss weighting is particularly effective when objectives have different spatial characteristics.

However, Token+Pixel exhibits an inverted scaling pattern: k-NN peaks at 32 experts then drops at 64 experts (69.3\%), a phenomenon not observed in Token+CLS. We attribute this to weaker gradient signals from pixel reconstruction, which may be insufficient to train 64 expert sets. MoE variants use the default pixel loss weight ($w_\text{pixel}\!=\!1.0$); the non-MoE baseline uses tuned weighting ($w_\text{pixel}\!=\!0.01$). While this 100$\times$ difference could partially explain the Token+Pixel gap, the comparison isolates the MoE contribution: dispatch weights provide implicit loss modulation that partially substitutes for manual weight tuning, and the weighted vs.\ unweighted comparison ($+$0.7\% at 32 experts) controls for this confound since both use $w_\text{pixel}\!=\!1.0$.

\textbf{Parameter cost.} Expert scaling increases parameters substantially: 86M (no MoE) $\to$ 116M (2 experts, +35\%) $\to$ 1.86B (64 experts, 21.6$\times$). The weighted vs.\ unweighted comparisons at matched expert counts (Table~\ref{tab:scaling}) isolate loss weighting from the parameter effect; a detailed mechanism isolation analysis is in supplementary Section~I. We view the 2-expert configuration as the practical recommendation (35\% parameter increase for 79.6\% linear probe), while the 64-expert scale demonstrates that dispatch-weight loss weighting improves representation quality at scale.

% \begin{table}[t]
% \centering
% \small
% \caption{\textbf{Expert scaling analysis.} Token+CLS scales smoothly to 64 experts (see also Figure~\ref{fig:expert_scaling}a for intermediate expert counts); Token+Pixel peaks at 32. All MoE models use sparse encoding, 300 epochs. W=dispatch-weight loss weighting. $\star$~Dense encoding (BEiT-style~\cite{bao2021beit}). $\dagger$~Tuned pixel loss weight ($w_\text{pixel}\!=\!0.01$).}
% \label{tab:scaling}
% \begin{tabular}{llccc}
% \toprule
% Model & Experts & W & k-NN@20 & Linear \\
% \midrule
% \multicolumn{5}{c}{\textit{Token + CLS}} \\
% \midrule
% MaskDistill~\cite{peng2022unified} (reprod.)$^\star$ & --- & --- & 75.6 & 75.7 \\
% No MoE & --- & --- & 75.7 & 76.2 \\
% ExPLoRe & 2 & --- & 74.1 & --- \\
% ExPLoRe & 2 & $\checkmark$ & 75.4 & 79.6 \\
% ExPLoRe & 64 & --- & 75.3 & --- \\
% \textbf{ExPLoRe} & \textbf{64} & $\checkmark$ & \textbf{76.2} & \textbf{80.6} \\
% \midrule
% \multicolumn{5}{c}{\textit{Token + Pixel}} \\
% \midrule
% No MoE$^{\dagger}$ & --- & --- & 71.5 & --- \\
% ExPLoRe & 32 & --- & 73.0 & --- \\
% \textbf{ExPLoRe} & \textbf{32} & $\checkmark$ & \textbf{73.7} & 79.4 \\
% ExPLoRe & 64 & $\checkmark$ & 69.3 & --- \\
% \bottomrule
% \end{tabular}
% \end{table}

\begin{table}[t]
\centering
\small
\setlength{\tabcolsep}{1.5pt}
\caption{\textbf{Expert scaling analysis.} Token+CLS scales smoothly to 64 experts (see also Figure~\ref{fig:expert_scaling}a); Token+Pixel peaks at 32. All MoE models use sparse encoding, 300 epochs. W=dispatch-weight loss weighting. $\star$~Dense encoding (BEiT-style~\cite{bao2021beit}). $\dagger$~Tuned pixel loss weight ($w_\text{pixel}\!=\!0.01$).}
\label{tab:scaling}

\begin{tabular}{lcccccc}
\toprule
\textbf{Token + CLS} &
MaskDistill$^\star$ & No MoE & 2 & 2+W & 64 & \textbf{64+W} \\
\midrule
k-NN@20 & 75.6 & 75.7 & 74.1 & 75.4 & 75.3 & \textbf{76.2} \\
Linear  & 75.7 & 76.2 & ---  & 79.6 & ---  & \textbf{80.6} \\
\bottomrule
\end{tabular}

\vspace{1pt}
\begin{tabular}{lcccc}
\toprule
\textbf{Token + Pixel} &
No MoE$^\dagger$ & 32 & \textbf{32+W} & 64+W \\
\midrule
k-NN@20 & 71.5 & 73.0 & \textbf{73.7} & 69.3 \\
Linear  & ---  & ---  & 79.4 & --- \\
\bottomrule
\end{tabular}
\end{table}

% Expert dispatch-weight visualization moved to introduction (Figure~\ref{fig:expert_dispatch})

%% Routing Dynamics (merged into Mechanism Validation)

\textbf{Entropy Regularization and Stability Transitions.}
We sweep entropy regularization weight $\lambda$ on the 2-expert Token+CLS configuration (Table~\ref{tab:ablation}). We use two diagnostics: the dispatch-weight coefficient of variation (CV) across patches, and the \emph{weighted-to-unweighted loss ratio} (MoE-weighted loss divided by the same loss computed with uniform weights; 1.0 indicates neutral routing, $\ll\!1.0$ indicates the router is redistributing weight away from difficult patches). A sharp transition exists between $\lambda=0.01$ (collapsed, 66.9\% k-NN) and $\lambda=0.1$ (stable, 75.3\%): CV jumps 32$\times$ and the loss ratio drops to 0.13, indicating deceptive loss minimization rather than meaningful specialization.

At $\lambda=0.5$ (optimal for 2 experts), k-NN reaches 75.5\% (+0.1\% over the default $\lambda=5.0$). Critically, optimal entropy weight is \emph{expert-count dependent}: while $\lambda=0.5$ helps 2 experts (+0.1\%), it degrades 64-expert performance ($-$0.2\% vs.\ $\lambda=5.0$). With 64 experts, inherent diversity from more expert parameters provides natural dispatch variation, and lower regularization leads to instability rather than improved contrast.

\textbf{Loss block selection.} We apply dispatch-weight loss weighting at the last MoE block (Block~11). Ablations over blocks \{5, 7, 9, 11\} show Block~11 performs best (75.4\% k-NN), with a narrow 0.2\% spread across blocks, confirming that the last block captures the most task-relevant routing patterns. Multi-block fusion (combining weights from blocks 7+9+11) underperforms single-block selection, suggesting that intermediate blocks add noise rather than complementary signal.

\textbf{CLS token specialization.} At Block~11 (the loss block), Expert~1 concentrates 91--95\% of its dispatch weight on the CLS token, effectively becoming a CLS specialist. This concentration emerges exclusively at the loss-coupled block; non-loss blocks show no such pattern. The phenomenon provides direct evidence that loss-coupling drives expert specialization: the router learns to route the CLS token heavily to the expert weighting CLS-aligned loss.

%% ─────────────────────────────────────────────────────────────
%% §4.6  Downstream Transfer Recipes
%% Story: MoE needs special treatment for downstream tasks
%% ─────────────────────────────────────────────────────────────
\subsection{Downstream Transfer Recipes}
\label{sec:recipe_results}

While MoE loss weighting improves pretraining representations, transferring MoE models to downstream tasks is non-trivial: standard finetuning can overwrite pretrained routing patterns, and extra expert parameters increase overfitting risk. We evaluate three complementary strategies: \textbf{Freeze Routing (FR)} freezes router parameters ($\mathbf{\Phi}$, $s$) to preserve pretrained patch-expert assignments~\cite{zoph2022stmoe}; \textbf{Freeze Attention (FA)} additionally freezes attention weights~\cite{zoph2022stmoe}; and \textbf{Expert Dropout (ExD)} applies dropout ($p\!=\!0.4$) to expert outputs~\cite{fedus2021switch}. Full rationale is in supplementary Section~J.

Table~\ref{tab:recipes} presents our recipe campaign on the Token+CLS 64-expert model, building each step on the previous best. Without any adaptation recipe, standard finetuning with unfrozen routing achieves 83.8\%, as the router overwrites pretrained routing patterns with task-specific shortcuts. Freeze Routing (FR) improves to 84.2\% (+0.4\%) by preserving pretrained patch-expert assignments. Adding Freeze Attention (FA) yields 84.3\%, a modest further gain. Adding Expert Dropout to FR provides the largest single gain (+0.7\%), yielding FR+ExD at 84.9\% by preventing over-reliance on dominant experts. The full \textbf{FR+FA+ExD} recipe reaches \textbf{85.3\%}, a +1.5\% improvement over vanilla MoE finetuning (83.8\%), exceeding the non-MoE baseline (84.8\%) by +0.5\%.

The recipe transfers across objective combinations and expert counts: the Token+Pixel 32-expert model improves from 83.6\% to 84.8\% with FR+FA+ExD (+1.2\%), exceeding the Token+Pixel non-MoE baseline (84.6\%). The 2-expert Token+CLS model with FR alone achieves 84.1\%, below the non-MoE baseline (84.8\%), indicating that the full recipe stack is needed to overcome the additional overfitting risk from expert parameters even at the 2-expert scale.

\begin{table}[t]
\centering
\small
\caption{\textbf{Downstream adaptation recipes.} FR=Freeze Routing, FA=Freeze Attention, ExD=Expert Dropout ($p\!=\!0.4$). Top: Token+CLS 64-expert recipe ablation. Bottom: cross-configuration transfer.}
\label{tab:recipes}
\begin{tabular}{lcccc}
\toprule
Configuration & FR & FA & ExD & Top-1 \\
\midrule
\multicolumn{5}{c}{\textit{Token+CLS 64-Expert}} \\
\midrule
Vanilla (unfrozen routing) & & & & 83.8 \\
FR only & $\checkmark$ & & & 84.2 \\
FR+FA & $\checkmark$ & $\checkmark$ & & 84.3 \\
FR+ExD & $\checkmark$ & & $\checkmark$ & 84.9 \\
\textbf{FR+FA+ExD} & $\checkmark$ & $\checkmark$ & $\checkmark$ & \textbf{85.3} \\
\midrule
\multicolumn{5}{c}{\textit{Cross-Configuration Transfer}} \\
\midrule
Token+CLS (no MoE) & --- & --- & --- & 84.8 \\
Token+CLS 2-expert (FR) & $\checkmark$ & & & 84.1 \\
Token+Pixel 32-expert (Vanilla) & & & & 83.6 \\
Token+Pixel 32-expert (FR+FA+ExD) & $\checkmark$ & $\checkmark$ & $\checkmark$ & 84.8 \\
Token+Pixel (no MoE) & --- & --- & --- & 84.6 \\
\bottomrule
\end{tabular}
\end{table}

%% ─────────────────────────────────────────────────────────────
%% §4.7  Semantic Segmentation
%% Story: Recipes close the gap — MoE+FR+FA+ExD exceeds baselines
%% ─────────────────────────────────────────────────────────────
\subsubsection{Semantic Segmentation}
\label{sec:semseg_results}

Table~\ref{tab:semseg} evaluates semantic segmentation on ADE20K for both objective combinations at 32 experts, enabling controlled comparison at a common scale (the optimal expert count for Token+Pixel, Table~\ref{tab:scaling}). MoE models present a well-known challenge for downstream transfer~\cite{zoph2022stmoe,fedus2021switch}: routing patterns optimized during pretraining can conflict with the uniform spatial coverage required for dense prediction. Without adaptation recipes, our MoE models underperform non-MoE baselines by 2.5--2.9 mIoU, consistent with prior findings that MoE finetuning is non-trivial~\cite{zoph2022stmoe}. Even in this default setting, dispatch-weight loss weighting helps modestly: +0.5 mIoU (Token+Pixel) and +0.6 mIoU (Token+CLS) over unweighted variants.

The adaptation recipes progressively close this gap. Freeze Routing (FR) alone recovers 0.7--1.4 mIoU by preserving pretrained routing patterns. The full FR+FA+ExD recipe fully closes the gap, with MoE models reaching parity or slight improvement over non-MoE baselines: Token+Pixel reaches \textbf{52.8} mIoU (+0.3 over non-MoE 52.5), and Token+CLS reaches \textbf{51.1} (+0.3 over non-MoE 50.8). This recovery arc, from MoE hurting dense prediction to matching baselines with proper adaptation, demonstrates that the recipes are essential for realizing MoE benefits beyond classification.

% \begin{table}[t]
% \centering
% \small
% \caption{\textbf{Semantic segmentation on ADE20K} (UperNet, 160K iters). Without recipes, MoE underperforms baselines; the full FR+FA+ExD recipe exceeds non-MoE for both objectives. W=dispatch-weight loss weighting.}
% \label{tab:semseg}
% \begin{tabular}{llcc}
% \toprule
% Model & Recipe & W & mIoU \\
% \midrule
% \multicolumn{4}{c}{\textit{Token + Pixel}} \\
% \midrule
% No MoE & Default & --- & 52.5 \\
% 32-expert & Default & --- & 49.6 \\
% 32-expert & Default & $\checkmark$ & 50.1 \\
% 32-expert & FR & $\checkmark$ & 50.8 \\
% \textbf{32-expert} & \textbf{FR+FA+ExD} & $\checkmark$ & \textbf{52.8} \\
% \midrule
% \multicolumn{4}{c}{\textit{Token + CLS}} \\
% \midrule
% No MoE & Default & --- & 50.8 \\
% 32-expert & Default & --- & 48.3 \\
% 32-expert & Default & $\checkmark$ & 48.9 \\
% 32-expert & FR & $\checkmark$ & 50.3 \\
% \textbf{32-expert} & \textbf{FR+FA+ExD} & $\checkmark$ & \textbf{51.1} \\
% \bottomrule
% \end{tabular}
% \end{table}

\begin{table}[t]
\centering
\small
\setlength{\tabcolsep}{1.3pt}
\caption{\textbf{Semantic segmentation on ADE20K} (UperNet, 160K iters, 32-expert). T+P=Token+Pixel, T+C=Token+CLS. Def=default finetuning (no recipe), W=dispatch-weight loss weighting, \textbf{All}=FR+FA+ExD.}
\label{tab:semseg}

\begin{minipage}[t]{0.45\linewidth}
\centering
\begin{tabular}{@{}lccccc@{}}
\toprule
\textbf{T+P} & No MoE & Def & Def+W & FR+W & \textbf{All+W} \\
\midrule
mIoU & 52.5 & 49.6 & 50.1 & 50.8 & \textbf{52.8} \\
\bottomrule
\end{tabular}
\end{minipage}\hfill
\begin{minipage}[t]{0.45\linewidth}
\centering
\begin{tabular}{@{}lccccc@{}}
\toprule
\textbf{T+C} & No MoE & Def & Def+W & FR+W & \textbf{All+W} \\
\midrule
mIoU & 50.8 & 48.3 & 48.9 & 50.3 & \textbf{51.1} \\
\bottomrule
\end{tabular}
\end{minipage}
\end{table}

%% ─────────────────────────────────────────────────────────────
%% §4.8  Comparison with Published Methods
%% Story: Cap-off — competitive position after building the case
%% ─────────────────────────────────────────────────────────────
\subsubsection{Comparison with Published Methods}
\label{sec:comparison}

Table~\ref{tab:comparison} compares ExPLoRe against published methods using ViT-Base encoders. Our 2-expert model (116M) achieves 79.6\% linear probe, +3.4\% over MaskDistill (76.2\%). Scaling to 64 experts reaches \textbf{80.6\%} linear probe, the highest among compared methods, with 85.3\% finetuning accuracy matching CAE~v2. Semantic segmentation with the full recipe reaches 52.8\% mIoU (Token+Pixel 32-expert, Table~\ref{tab:semseg}), competitive with BEiT~v2 (52.7\%) and MILAN (52.7\%). MaskDistill's 53.8\% segmentation is from the original paper~\cite{peng2022unified}; our controlled baselines in Table~\ref{tab:semseg} provide the direct comparison.

Although the 64-expert model has 1.86B parameters, Soft-MoE with one slot per expert decouples parameters from inference cost: each MoE layer routes all tokens into only $E$ expert slots, so the expert MLPs process $E$ rather than $N$ inputs and only the routing projections scale with $E$. Both ExPLoRe configurations therefore use \emph{fewer} inference GFLOPs than every ViT-B/16 comparator (11.93 and 13.86 vs.\ 17.45; Table~\ref{tab:comparison}). Per-FLOP, ExPLoRe is favorably positioned: E=2 trades $-32\%$ cost for a $-0.9\%$ linear-probe gap to CAE~v2, and E=64 matches CAE~v2 at $-21\%$ cost (cost--quality plot in the supplementary).

\begin{table}[t]
\centering
\small
\caption{\textbf{Comparison with published methods on ImageNet-1K.} All use ViT-Base encoders. GFLOPs=inference cost (analytical MACs at $N\!=\!197$ tokens, cross-validated against fvcore within 1.1\%). Lin.=linear probe top-1, FT=finetuning top-1, Seg.=ADE20K mIoU. ``---'' = not reported. $\ast$~Our reproduction with sparse Token+CLS encoding; Lin./FT are ours, Seg.\ from~\cite{peng2022unified}. $\dagger$~400 epochs. $\ddagger$~YFCC-15M pretrained. $\S$~Token+Pixel 32-exp with FR+FA+ExD (Section~\ref{sec:recipe_results}).}
\label{tab:comparison}
\begin{tabular}{lccccccc}
\toprule
Method & Teacher & Params & GFLOPs & Ep. & Lin. & FT & Seg. \\
\midrule
MAE~\cite{he2022masked} & None & 86M & 17.45 & 1600 & 68.0 & 83.6 & 48.1 \\
SdAE~\cite{chen2022sdae} & Self & 86M & 17.45 & 300 & --- & 84.1 & 48.6 \\
BootMAE~\cite{dong2022bootstrapped} & Self & 86M & 17.45 & 800 & --- & 83.6 & --- \\
\midrule
BEiT~\cite{bao2021beit} & DALL-E & 86M & 17.45 & 800 & 56.7 & 83.0 & 45.6 \\
MVP~\cite{wei2022mvp} & CLIP-B & 86M & 17.45 & 300 & 75.4 & 84.4 & 52.4 \\
MaskDistill$^\ast$ (repr.)~\cite{peng2022unified} & CLIP-B & 86M & 17.45 & 300 & 76.2 & 84.8 & \textbf{53.8} \\
BEiT v2~\cite{peng2022beit} & CLIP-B & 86M & 17.45 & 300 & 80.1 & 85.0 & 52.7 \\
MILAN$^\dagger$~\cite{hou2022milan} & CLIP-B & 86M & 17.45 & 400 & 79.9 & \textbf{85.4} & 52.7 \\
CAE v2~\cite{zhang2024caev2} & CLIP-B & 86M & 17.45 & 300 & 80.5 & 85.3 & 52.9 \\
MaskCLIP$^\ddagger$~\cite{dong2023maskclip} & CLIP-B & 86M & 17.45 & 25 & 73.7 & 83.6 & 50.5 \\
\midrule
ExPLoRe (ours, 2-exp) & CLIP-B & 116M & \textbf{11.93} & 300 & 79.6 & 84.1 & --- \\
ExPLoRe (ours, 64-exp) & CLIP-B & 1.86B & \textbf{13.86} & 300 & \textbf{80.6} & 85.3 & 52.8$^\S$ \\
\bottomrule
\end{tabular}
\end{table}

%% ─────────────────────────────────────────────────────────────
%% §4.9  Robustness and additional probing protocols
%% ─────────────────────────────────────────────────────────────
\subsubsection{Robustness and Additional Probing Protocols}
\label{sec:robustness}

\noindent\textbf{Seed and mask-ratio robustness.} Across two additional seeds and a 50\% mask ratio, the 2-expert Token+CLS model varies tightly: standard deviation 0.078 (k-NN), 0.037 (linear probe), 0.039 (Efficient Probing), with mask=50\% within this band (40\% is the established block-masking optimum~\cite{bao2021beit,peng2022unified}). Our mechanism-isolation deltas ($+1.3\%$ weighted vs.\ unweighted, $-1.6\%$ detach) compare configurations sharing pretraining randomness and are not single-run artifacts; full per-seed numbers are in the supplementary.

\noindent\textbf{Efficient Probing and fine-grained transfer.} Under Efficient Probing (EP)~\cite{psomas2026attention}, a lightweight attentive protocol recovering patch-level information that CLS-aggregated probes under-measure, the 2-expert model reaches \textbf{80.19\%} vs.\ \textbf{78.77\%} for the non-MoE baseline ($+1.41\%$, $\sim\!3\times$ the k-NN gain on the same backbones). On fine-grained Food-101~\cite{bossard2014food}, EP gives \textbf{89.55\%} vs.\ \textbf{87.68\%} ($+1.87\%$), indicating the routing emphasis transfers beyond ImageNet.

\section{Conclusion}
\label{sec:conclusion}

We presented ExPLoRe, a framework that repurposes Soft-MoE dispatch weights as learned per-patch loss coefficients for multi-objective masked image modeling. Three findings emerge from our study.
\textbf{(1)~Loss-coupling drives specialization:} detaching this connection degrades performance by 1.6\%, and MoE without loss-coupling underperforms the non-MoE baseline (74.1\% vs.\ 75.7\%), confirming the mechanism is essential. While absolute k-NN gains are modest (+0.5\% for Token+CLS, +2.2\% for Token+Pixel), the +4.4\% linear probe improvement suggests loss-coupled routing reshapes the feature space in ways not fully captured by nearest-neighbor evaluation.
\textbf{(2)~Routing dynamics exhibit sharp transitions:} entropy regularization governs a sharp transition between stable routing and catastrophic collapse, with the optimal weight being expert-count dependent ($\lambda\!=\!0.5$ for 2 experts, $\lambda\!=\!5.0$ for 64).
\textbf{(3)~MoE transfer requires dedicated recipes:} our FR+FA+ExD recipe improves MoE finetuning by +1.5\% over vanilla MoE (83.8\% $\to$ 85.3\%), exceeding the non-MoE baseline (84.8\%). Without recipes, MoE models underperform on segmentation by 2.5--2.9 mIoU; the full recipe closes this gap (+0.3 mIoU for both objective combinations).

\textbf{Limitations and Future Work.}
Our experiments use ViT-Base with up to 64 experts (GPU-memory bounded) and a frozen CLIP teacher, though the mechanism is teacher-agnostic and applies to any multi-objective MIM framework. Natural extensions include a \emph{dedicated loss-weighting router} separate from the feature router (enabling top-$k$ routing across all experts per loss), \emph{multiple slots per expert} ($S\!>\!1$), \emph{learned block aggregation} of dispatch weights across MoE layers, and scaling to larger models and longer schedules.

% ---- Bibliography ----
\bibliographystyle{splncs04}
\bibliography{references}

% ---- Supplementary material as appendix (arXiv combined version) ----
\clearpage
\appendix
\renewcommand{\thetable}{A\arabic{table}}
\renewcommand{\thefigure}{A\arabic{figure}}
\renewcommand{\theequation}{A\arabic{equation}}
\setcounter{table}{0}
\setcounter{figure}{0}
\setcounter{equation}{0}
\section{Training and Evaluation Hyperparameters}
\label{sec:suppl_hyperparams}

\subsection{Pretraining Configuration}

Table~\ref{tab:suppl_pretrain} summarizes the pretraining hyperparameters shared across all model configurations. All experiments use identical optimization settings to ensure fair comparison; only the loss objectives and MoE configuration vary between runs.

\begin{table}[ht]
\centering
\small
\caption{\textbf{Pretraining hyperparameters.} Shared across all configurations.}
\label{tab:suppl_pretrain}
\begin{tabular}{ll}
\toprule
Parameter & Value \\
\midrule
\multicolumn{2}{c}{\textit{Architecture}} \\
\midrule
Encoder & ViT-Base/16 (86M params) \\
Teacher & Frozen CLIP ViT-B/16 \\
Decoder & 8 blocks, dim 512 \\
Position embeddings (encoder) & Absolute (learnable) \\
Position embeddings (decoder) & Learnable \\
\midrule
\multicolumn{2}{c}{\textit{Masking}} \\
\midrule
Strategy & Block masking \\
Mask ratio & 40\% \\
Patch shuffling & Enabled (sparse mode) \\
\midrule
\multicolumn{2}{c}{\textit{Optimization}} \\
\midrule
Optimizer & AdamW ($\beta_1\!=\!0.9$, $\beta_2\!=\!0.95$) \\
Weight decay & 0.05 \\
Peak learning rate & $1.5 \times 10^{-3}$ \\
Minimum learning rate & $1 \times 10^{-6}$ \\
Schedule & Cosine decay \\
Warmup epochs & 40 \\
Total epochs & 300 \\
Batch size & 4096 (across 4$\times$ H100 GPUs) \\
Gradient clipping & 3.0 \\
Mixed precision & FP16 \\
Seed & 42 \\
\bottomrule
\end{tabular}
\end{table}

\subsection{MoE Configuration Details}

Table~\ref{tab:suppl_moe} details the MoE-specific parameters for each model configuration evaluated in the main paper.

\begin{table}[ht]
\centering
\small
\caption{\textbf{MoE configuration details.} Parameters specific to each ExPLoRe variant.}
\label{tab:suppl_moe}
\begin{tabular}{lcc}
\toprule
Parameter & Token+CLS & Token+Pixel \\
\midrule
\multicolumn{3}{c}{\textit{Objectives}} \\
\midrule
Token distillation & $\checkmark$ & $\checkmark$ \\
CLS alignment ($w\!=\!0.4$) & $\checkmark$ & -- \\
Pixel reconstruction & -- & $\checkmark$ \\
\midrule
\multicolumn{3}{c}{\textit{MoE Architecture}} \\
\midrule
MoE placement & Alternating (blocks 1,3,5,7,9,11) & Same \\
Slots per expert & 1 & 1 \\
Routing type & Soft (fully differentiable) & Same \\
Scale init & 1.0 (learned) & Same \\
Expert init & Kaiming uniform & Same \\
\midrule
\multicolumn{3}{c}{\textit{Regularization}} \\
\midrule
Entropy weight (dispatch) & 5.0 (default) & 5.0 \\
Loss weighting block & Block 11 (last MoE) & Same \\
Per-image normalization & $\checkmark$ & $\checkmark$ \\
Loss weighting method & Dispatch weights & Dispatch weights \\
\bottomrule
\end{tabular}
\end{table}

\subsection{Downstream Evaluation Protocols}

\textbf{k-NN Classification.} We use $k\!=\!20$ nearest neighbors with cosine similarity on [CLS] token representations. Features are extracted from the last encoder block without any learned components, providing a training-free measure of representation quality. We evaluate checkpoints at epochs \{100, 150, 180, 200, 220, 240, 250, 260, 270, 280, 290, 295, 300\} and report the best.

\textbf{Linear Probing.} A single linear layer is trained on frozen [CLS] token features for 100 epochs with batch size 2048, learning rate $2 \times 10^{-3}$ with cosine decay, 10-epoch warmup, label smoothing 0.1, mixup 0.8, cutmix 1.0, and RandAugment (9, 0.5).

\textbf{ImageNet Finetuning.} Full model finetuning for 100 epochs with layer-wise learning rate decay (base $5 \times 10^{-4}$, decay factor 0.65), batch size 1024, AdamW with weight decay 0.05, warmup 5 epochs, mixup 0.8, cutmix 1.0, drop path 0.1. MoE models use the FR+FA+ExD recipe (Section~4.4 of the main paper) unless noted otherwise.

\textbf{Semantic Segmentation.} UperNet~\cite{xiao2018unified} on ADE20K with 160K iterations, learning rate $8 \times 10^{-5}$, layer-wise decay 0.85, batch size 16. MoE models freeze routing parameters and use expert dropout ($p\!=\!0.4$).

\section{GradNorm Hyperparameter Search Details}
\label{sec:suppl_gradnorm}

Table~\ref{tab:suppl_gradnorm} presents the complete GradNorm hyperparameter search across both objective combinations. For Token+CLS (dense encoding), we explored $\alpha \in \{0.12, 0.5, 1.0, 1.5\}$ and learning rates $\in [10^{-4}, 0.15]$ across 5 configurations; the best result (72.8\% k-NN) substantially underperforms the static baseline (74.2\%). For Token+Pixel (dense encoding), we tested 12 configurations spanning $\alpha \in \{0.0, 0.5, 1.0\}$ and learning rates $\in [0.00125, 0.025]$; all configurations exhibited training instability and collapsed to $\leq$32.5\% k-NN, far below the non-MoE baseline (71.5\%). The consistent failure across diverse hyperparameters confirms that GradNorm's global gradient-based balancing is fundamentally incompatible with spatially heterogeneous MIM objectives, motivating the per-patch approach of ExPLoRe.

\begin{table}[ht]
\centering
\small
\caption{\textbf{Complete GradNorm hyperparameter search.} Dense encoding throughout. All Token+CLS configurations underperform the static baseline (74.2\% k-NN). All Token+Pixel configurations collapse. Representative results from 12 Token+Pixel configurations shown.}
\label{tab:suppl_gradnorm}
\begin{tabular}{ccccc}
\toprule
$\alpha$ & Learning Rate & Objectives & k-NN@20 & Status \\
\midrule
\multicolumn{5}{c}{\textit{Token + CLS (baseline: 74.2\%)}} \\
\midrule
0.12 & 0.15 & Token + CLS & 72.8 & Best \\
0.12 & 0.0001 & Token + CLS & 69.5 & \\
1.0 & 0.025 & Token + CLS & 68.7 & \\
1.5 & 0.025 & Token + CLS & 68.3 & \\
0.5 & 0.00015 & Token + CLS & 66.4 & Worst \\
\midrule
\multicolumn{5}{c}{\textit{Token + Pixel (baseline: 71.5\%)}} \\
\midrule
1.0 & 0.025 & Token + Pixel & 32.5 & Best \\
0.5 & 0.025 & Token + Pixel & 30.8 & \\
0.0 & 0.00125 & Token + Pixel & 28.3 & \\
\bottomrule
\end{tabular}
\end{table}

\section{Entropy Regularization Analysis}
\label{sec:suppl_entropy}

We conducted a grid search to identify optimal entropy regularization weights for the 2-expert Token+Pixel configuration (dense encoding). For Expert~0 (token distillation) we explored $\lambda_0 \in \{0.1, 0.5, 1.0, 2.0, 3.0, 5.0\}$, and for Expert~1 (pixel reconstruction) we explored $\lambda_1 \in \{5.0, 10.0, 15.0, 20.0, 24.0, 30.0\}$.

The optimal configuration uses asymmetric weights: $\lambda_0\!=\!5.0$, $\lambda_1\!=\!24.0$. The 4.8$\times$ asymmetry compensates for weaker gradient signals from pixel reconstruction compared to semantic distillation. Configurations with $\lambda_1 < 20.0$ consistently exhibited expert collapse, with Expert~1 receiving $<$10\% of tokens. For the Token+CLS configuration (sparse encoding), symmetric weights ($\lambda\!=\!5.0$ for both experts) are sufficient since both objectives involve semantic features.

\section{Expert Routing Visualizations}
\label{sec:suppl_routing_viz}

This section provides visual and quantitative evidence that loss-coupled dispatch weights develop content-dependent expert specialization. We compare the coupled model (ExPLoRe with $\lambda_\text{entropy}\!=\!0.5$, the best 2-expert configuration from Section~4.2) against the \emph{detach} ablation (trained with identical $\lambda_\text{entropy}\!=\!0.5$), which uses identical dispatch weights for the MoE forward pass but detaches gradients before loss weighting, removing the loss-coupling signal.

\subsection{Dispatch Weight Heatmaps}

\begin{figure}[!ht]
\centering
\includegraphics[width=\textwidth]{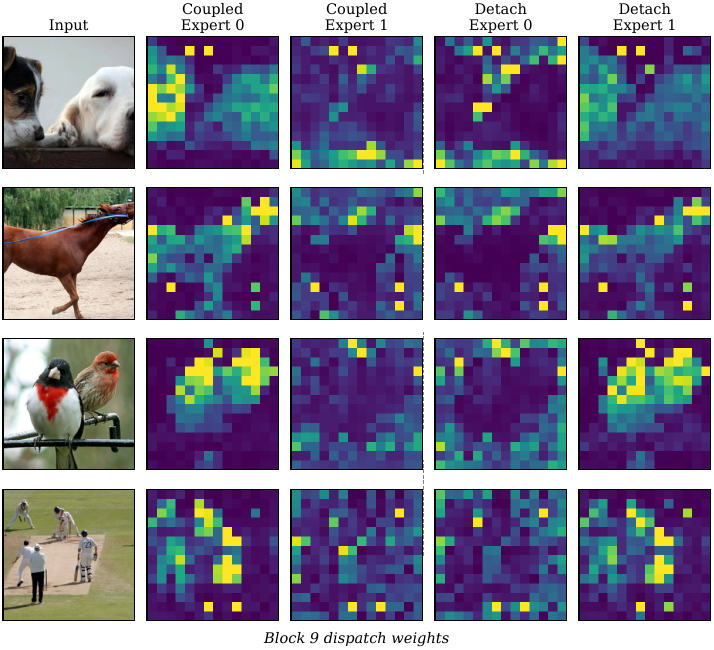}
\caption{\textbf{Dispatch weight heatmaps.} Each row shows a different input image. \emph{Left pair:} Coupled (ExPLoRe) Expert~0 and Expert~1 dispatch weights. \emph{Right pair:} Detach ablation. The coupled model produces spatially coherent, content-dependent routing; the detach model shows more scattered patterns. Dashed line separates the two models. Per-model color normalization.}
\label{fig:dispatch_heatmaps}
\end{figure}

Figure~\ref{fig:dispatch_heatmaps} visualizes per-patch dispatch weights for four natural images. In the coupled model, Expert~0 shows spatially coherent activation regions aligned with salient image content (\eg, the dog's body, the horse, the bird), while Expert~1 exhibits complementary patterns. In the detach model, both experts show more scattered, less content-aligned activation patterns. Per-model normalization is applied so that each coupled/detach pair uses its full color range, enabling fair comparison of spatial structure rather than absolute magnitude.

\subsection{Expert Cluster Separation Across Blocks}

To quantify routing specialization, we compute the silhouette coefficient~\cite{rousseeuw1987silhouettes} (ranging from $-1$ to $+1$, where higher values indicate better-separated clusters) at each MoE block, using per-token input embeddings as features and $\text{argmax}(\text{dispatch weights})$ as cluster labels.

\begin{figure}[!ht]
\centering
\includegraphics[width=0.75\textwidth]{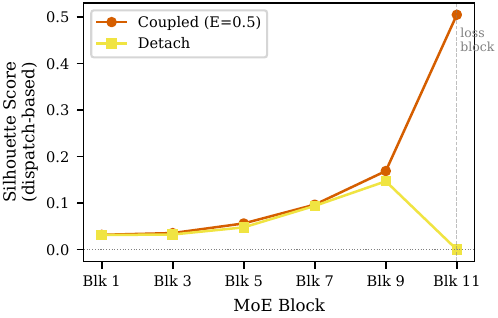}
\caption{\textbf{Silhouette coefficient across MoE blocks.} Dispatch-based cluster assignments on 200 ImageNet validation images (50K tokens subsampled). The coupled model develops strong expert specialization at Block~11 (the loss block), while the detach ablation collapses to zero specialization at this block. Early blocks show similar routing structure in both models.}
\label{fig:silhouette}
\end{figure}

Figure~\ref{fig:silhouette} shows that both models exhibit similar, modest specialization at early blocks (1--5), consistent with generic feature extraction. The models diverge at later blocks: at Block~9, the coupled model reaches a silhouette of 0.17 vs.\ 0.15 for detach. At Block~11 (the loss block), the divergence is dramatic: the coupled model achieves 0.505 while the detach model collapses to 0.0 (all tokens assigned to a single expert). This confirms that loss-coupling drives expert specialization \emph{specifically at the loss block}. Without gradient flow through dispatch weights, the entropy regularization pushes the router toward uniformity with no counteracting signal, eliminating specialization entirely.

\subsection{Dispatch Weight--Loss Correlation}

\begin{figure}[!ht]
\centering
\includegraphics[width=\textwidth]{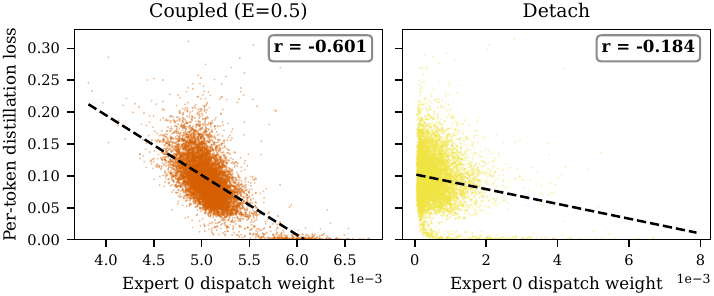}
\caption{\textbf{Dispatch weight vs.\ per-token distillation loss at Block~11.} Each point is one token from 10K subsampled tokens across ImageNet validation images. \emph{Left:} Coupled model shows strong negative correlation ($r\!=\!{-}0.601$), indicating the router upweights patches where it reduces loss. \emph{Right:} Detach ablation shows weak correlation ($r\!=\!{-}0.191$). Dashed line: linear regression (clipped at $y\!=\!0$).}
\label{fig:scatter_correlation}
\end{figure}

To understand \emph{how} the coupled router specializes, we examine the relationship between Expert~0's dispatch weight and per-token distillation loss at Block~11. Figure~\ref{fig:scatter_correlation} shows scatter plots for both models. In the coupled model, there is a clear negative Pearson correlation ($r\!=\!{-}0.601$): patches receiving higher Expert~0 dispatch weight tend to have lower distillation loss, indicating that the router learns to upweight patches where it can reduce loss most effectively. In the detach model, this correlation nearly vanishes ($r\!=\!{-}0.191$), confirming that loss-coupling is necessary for the router to develop loss-relevant specialization.

\section{Encoding and Decoder Details}
\label{sec:suppl_encoding}

This section provides the full encoding equations summarized in Section~3.1 of the main paper, along with decoder architecture details.

\textbf{Dense Encoding (BEiT-style).} Following BEiT~\cite{bao2021beit}, masked patches are replaced with learnable mask tokens $\mathbf{e}_\text{mask} \in \mathbb{R}^d$, processing the complete sequence of $N$ patches:
\begin{equation}
\mathbf{x}_\text{dense} = \mathbf{x}_\text{vis} \odot (1 - \mathbf{m}) + \mathbf{e}_\text{mask} \odot \mathbf{m}
\end{equation}
where $\odot$ denotes element-wise multiplication. Dense encoding yields dispatch weights for all patches including masked ones, enabling MoE-weighted pixel reconstruction.

\textbf{Sparse Encoding (MAE-style).} Following MAE~\cite{he2022masked}, only visible patches are processed, reducing computation by factor $(1 - r)$ where $r$ is the mask ratio. Position embeddings are added \emph{before} masking to preserve spatial information:
\begin{equation}
\mathbf{x}_\text{pos} = \mathbf{x} + \mathbf{p}_\text{emb}
\end{equation}
where $\mathbf{p}_\text{emb} \in \mathbb{R}^{N \times d}$ are position embeddings. This pre-masking addition is crucial: once patches are masked and shuffled, their original spatial positions would be lost without embedded positional information. To prevent the model from exploiting positional regularities in the visible subset, patches undergo random shuffling:
\begin{equation}
\pi = \text{randperm}(N), \quad \mathbf{x}_\text{shuffle} = \mathbf{x}_\text{pos}[\pi]
\end{equation}
\begin{equation}
\mathbf{x}_\text{sparse} = \mathbf{x}_\text{shuffle}[\neg\mathbf{m}[\pi]]
\end{equation}
Without shuffling, the encoder could learn spurious correlations (\eg, ``patches at positions 0--49 are always visible for 75\% masking''). The inverse permutation $\pi^{-1}$ is stored for decoder reconstruction.

\textbf{Loss Application by Encoding Mode:}
\begin{itemize}
    \item \textbf{Dense}: Token distillation on masked positions $\mathcal{L}_\text{token}(\hat{\mathbf{t}}[\mathbf{m}], \mathbf{t}[\mathbf{m}])$, pixel reconstruction on masked positions $\mathcal{L}_\text{pixel}(\hat{\mathbf{x}}[\mathbf{m}], \mathbf{x}[\mathbf{m}])$
    \item \textbf{Sparse}: Token distillation on visible positions $\mathcal{L}_\text{token}(\hat{\mathbf{t}}[\neg\mathbf{m}], \mathbf{t}[\neg\mathbf{m}])$, pixel reconstruction on masked positions $\mathcal{L}_\text{pixel}(\hat{\mathbf{x}}[\mathbf{m}], \mathbf{x}[\mathbf{m}])$
\end{itemize}
The encoding mode determines which patches receive MoE dispatch weights and thus which losses can be MoE-weighted.

\textbf{Decoder Architecture.} The decoder is a lightweight 8-block transformer with embedding dimension 512 and 16 attention heads. In sparse mode, the decoder receives the encoder's visible-patch representations concatenated with learnable mask tokens at the masked positions, using the stored inverse permutation $\pi^{-1}$ to restore spatial ordering. Learnable position embeddings are added before decoding. A linear projection maps decoder outputs to the pixel target dimension ($16 \times 16 \times 3 = 768$) for reconstruction. The decoder is discarded after pretraining and is not used during downstream evaluation.

\section{Notation Reference}
\label{sec:suppl_notation}

Table~\ref{tab:suppl_notation} provides a compact reference for the notation used throughout the paper.

\begin{table}[ht]
\centering
\small
\caption{\textbf{Notation reference.}}
\label{tab:suppl_notation}
\begin{tabular}{ll}
\toprule
Symbol & Meaning \\
\midrule
$\mathbf{X} \in \mathbb{R}^{B \times N \times d}$ & Patch representations (batch, patches, dimension) \\
$\mathbf{\Phi} \in \mathbb{R}^{d \times E}$ & Expert routing parameters \\
$s$ & Learned routing scale parameter \\
$E$ & Number of experts \\
$\mathbf{L} \in \mathbb{R}^{B \times N \times E}$ & Routing logits (patch-expert affinity) \\
$\mathbf{D} \in \mathbb{R}^{B \times N \times E}$ & Dispatch weights (softmax over patches) \\
$\mathbf{C} \in \mathbb{R}^{B \times N \times E}$ & Combine weights (softmax over experts) \\
$\mathcal{V}$, $\mathcal{M}$ & Visible and masked patch sets \\
$\pi$, $\pi^{-1}$ & Shuffle permutation and its inverse \\
$\lambda$, $\lambda_e$ & Entropy regularization weight (global / per-expert) \\
$H_e$ & Per-expert dispatch entropy \\
$w_\text{cls}$, $w_\text{pixel}$ & CLS alignment / pixel reconstruction loss weights \\
\bottomrule
\end{tabular}
\end{table}

\section{Detailed Soft-MoE Routing Derivation}
\label{sec:suppl_softmoe_derivation}

This section provides the full step-by-step derivation of the Soft-MoE routing mechanism summarized in Section~3.2 of the main paper. We follow the formulation of Puigcerver et al.~\cite{softmoe}.

For a given MoE layer, we maintain $E$ learnable expert parameters $\mathbf{\Phi} \in \mathbb{R}^{d \times E}$ that define what type of content each expert responds to, along with $E$ expert networks each implemented as a standard MLP: $\text{MLP}(x) = W_2 \cdot \text{GELU}(W_1 \cdot x)$, where $W_1 \in \mathbb{R}^{d \times 4d}$ and $W_2 \in \mathbb{R}^{4d \times d}$. Each expert has independent parameters, allowing specialization through training. A learned scalar $s$ controls the sharpness of the routing distribution, similar to temperature in softmax.

The routing mechanism operates as follows. Given patch representations $\mathbf{X} \in \mathbb{R}^{B \times N \times d}$ from the attention layer:

\textbf{(1) Logits} measure patch-expert affinity through normalized similarity:
\begin{equation}
\mathbf{L} = s \cdot \frac{\mathbf{X}}{\|\mathbf{X}\|_2} \cdot \frac{\mathbf{\Phi}}{\|\mathbf{\Phi}\|_2}
\end{equation}
where L2 normalization stabilizes training and prevents logit saturation.

\textbf{(2) Dispatch weights} via softmax over the patch dimension aggregate patches to experts:
\begin{equation}
\mathbf{D} = \text{softmax}_{\text{dim}=1}(\mathbf{L}) \in \mathbb{R}^{B \times N \times E}
\end{equation}
where $\sum_{n} \mathbf{D}_{b,n,e} = 1$ for each expert $e$. These weights represent how much each patch contributes to each expert and are used for patch-level loss weighting.

\textbf{(3) Combine weights} via softmax over the expert dimension determine output composition:
\begin{equation}
\mathbf{C} = \text{softmax}_{\text{dim}=2}(\mathbf{L}) \in \mathbb{R}^{B \times N \times E}
\end{equation}
where $\sum_{e} \mathbf{C}_{b,n,e} = 1$ for each patch $n$. Dispatch and combine correspond to the input-to-slot and slot-to-output transformations in Soft-MoE; these are analogous to ``routing'' or ``gating'' weights in sparse MoE literature~\cite{shazeer2017outrageously,fedus2021switch}.

\textbf{(4) Aggregation} transforms patch representations into expert-specific inputs using dispatch weights:
\begin{equation}
\mathbf{S}_\text{in} = \mathbf{D}^\top \mathbf{X} \in \mathbb{R}^{B \times E \times d}
\end{equation}

\textbf{(5) Each expert} processes its inputs: $\mathbf{S}_\text{out}^{(e)} = \text{Expert}_e(\mathbf{S}_\text{in}^{(e)})$ for $e = 0, \ldots, E-1$.

\textbf{(6) Outputs are combined} back to per-patch representations:
\begin{equation}
\mathbf{Y} = \mathbf{C} \cdot \mathbf{S}_\text{out} \in \mathbb{R}^{B \times N \times d}
\end{equation}

\textbf{Parameter and Memory Cost.} Each MoE block replaces one MLP ($2 \times d \times 4d = 2 \times 768 \times 3072 \approx 4.7$M parameters) with $E$ independent expert MLPs, adding $(E-1) \times 4.7$M parameters per block across 6 alternating blocks. Total model parameters: 86M (no MoE), 116M ($E\!=\!2$, +35\%), 537M ($E\!=\!16$), 993M ($E\!=\!32$), 1.86B ($E\!=\!64$, 21.6$\times$). Peak GPU memory scales linearly with $E$ due to expert parameters and activations; $E\!=\!64$ requires approximately 72\,GB per GPU with batch size 1024 per GPU (FP16), fitting within 80\,GB H100 HBM3. Routing overhead (logits, dispatch, combine weights) is negligible: $3 \times B \times N \times E$ floats per MoE block.

\section{Extended Expert Scaling Discussion}
\label{sec:suppl_expert_scaling}

This section provides detailed analysis of expert behavior when the number of experts $E$ exceeds the number of loss-weighted objectives $K$, complementing the scaling results in Section~4.3 of the main paper.

\textbf{Two experts ($E\!=\!2$, $K\!=\!1$).} Expert~0 directly modulates the token loss through dispatch weights (loss-coupling). Expert~1 specializes through complementary gradient pressure: the shared logit matrix means that when Expert~0's dispatch weights are pushed toward foreground patches by loss gradients, Expert~1's weights shift toward the complementary set. Empirically, this manifests as CLS concentration (91--95\% dispatch weight on the CLS token at Block~11; Section~4.2 of the main paper).

\textbf{Many experts ($E\!=\!64$, $K\!=\!1$).} Only one expert receives direct loss-coupling; the remaining 63 provide capacity through standard MoE forward-pass gradients and entropy regularization. Even at this scale, dispatch-weight loss weighting provides a consistent benefit (+0.9\% k-NN over the unweighted variant), though the relative gain decreases from +1.3\% (2 experts) to +0.9\% (64 experts) as inherent diversity from more expert parameters provides partial dispatch variation independently of loss-coupling.

\section{Extended Mechanism Isolation Analysis}
\label{sec:suppl_mechanism_isolation}

This section details how the evaluation structure in Table~3 of the main paper isolates the loss-weighting mechanism from the parameter effect at each expert scale.

The isolation relies on three complementary comparisons:

\textbf{First}, comparing non-MoE to MoE quantifies the \emph{combined} effect of routing capacity and loss weighting. For Token+CLS: 75.7\% (no MoE) $\to$ 75.4\% (2-expert weighted) $\to$ 76.2\% (64-expert weighted).

\textbf{Second}, comparing weighted vs.\ unweighted variants \emph{at the same expert count} (identical parameters) isolates loss weighting: +1.3\% at 2 experts (74.1\% $\to$ 75.4\%), +0.9\% at 64 experts (75.3\% $\to$ 76.2\%).

\textbf{Third}, the unweighted 2-expert model (74.1\%) \emph{underperforms} the non-MoE baseline (75.7\%) despite having 35\% more parameters (116M vs.\ 86M), demonstrating that parameters alone do not explain the gains; the loss-coupling mechanism is essential.

The 2-expert configuration offers the most efficient tradeoff: 35\% parameter increase for 79.6\% linear probe (vs.\ 80.6\% at 64 experts with 21.6$\times$ parameters). We view the 64-expert scale as demonstrating that dispatch-weight loss weighting improves representation quality at scale, while the 2-expert configuration is the practical recommendation.

\section{Downstream Adaptation Strategies}
\label{sec:suppl_downstream_recipes}

While MoE-based loss weighting improves pretraining representations (as measured by k-NN and linear probe), transferring MoE models to downstream tasks presents unique challenges. Standard finetuning can disrupt the routing patterns learned during pretraining, and the additional expert parameters increase overfitting risk. Three complementary adaptation strategies address these challenges:

\textbf{Freeze Routing (FR).} Following ST-MoE~\cite{zoph2022stmoe}, we freeze all routing parameters ($\mathbf{\Phi}$, $s$) during finetuning, preserving pretrained patch-expert assignments while allowing expert MLP parameters to adapt. Without freezing, finetuning overwrites routing patterns with task-specific shortcuts.

\textbf{Expert Dropout (ExD).} Following Switch Transformer~\cite{fedus2021switch}, we apply dropout ($p\!=\!0.4$) to expert outputs during finetuning, preventing over-reliance on dominant experts and encouraging redundancy across the expert ensemble. This is analogous to standard dropout but operates at the expert granularity rather than the neuron level.

\textbf{Freeze Attention (FA).} Following ST-MoE~\cite{zoph2022stmoe}, we freeze all attention parameters (QKV and output projections), finetuning only MLP/expert parameters, layer norms, and the classification head. Pretrained attention patterns encode transferable visual structure.

These compose naturally: the best recipe \textbf{FR+FA+ExD} freezes routing and attention to preserve pretrained representations while expert dropout regularizes during task-specific adaptation. The recipe ablation and cross-configuration transfer results are in Section~4.4 of the main paper.

\section{Inference Cost and Cost--Quality Trade-off}
\label{sec:suppl_flops}

\begin{figure}[h]
\centering
\includegraphics[width=0.7\textwidth]{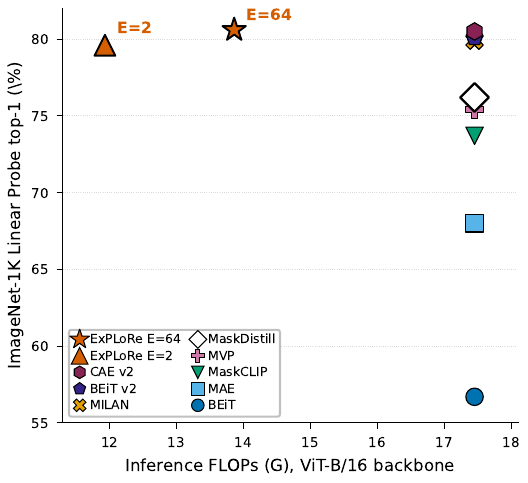}
\caption{\textbf{Cost--quality trade-off on ImageNet-1K.} Inference GFLOPs (ViT-B/16 backbone, $N\!=\!197$ tokens) vs.\ linear-probe top-1 accuracy. ExPLoRe E=2 (triangle) and E=64 (star) lie to the left of all compared methods, which cluster at 17.45 GFLOPs.}
\label{fig:costquality}
\end{figure}

Table~6 of the main paper reports inference GFLOPs as analytical MACs at $N\!=\!197$ tokens, cross-validated against fvcore within 1.1\%. Because Soft-MoE with one slot per expert routes all tokens into only $E$ expert slots, the expert MLPs process $E$ rather than $N$ inputs, so parameter count and inference cost decouple: both ExPLoRe configurations use fewer GFLOPs than every ViT-B/16 comparator (11.93 at $E\!=\!2$, 13.86 at $E\!=\!64$, vs.\ 17.45). Figure~\ref{fig:costquality} plots the resulting cost--quality trade-off: ExPLoRe sits to the upper-left of the comparator cluster, achieving comparable or higher linear-probe accuracy at lower inference cost. This directly addresses whether the mechanism provides value beyond simply adding parameters: on a per-FLOP basis it does.

\section{Seed and Mask-Ratio Robustness}
\label{sec:suppl_robustness}

To assess statistical reliability, we re-trained the 2-expert Token+CLS model under two additional random seeds and a higher 50\% mask ratio, evaluating each under k-NN, linear probe (LP), and Efficient Probing (EP). Table~\ref{tab:robust} reports the results: variation is tight across all three protocols (standard deviation 0.078, 0.037, 0.039 respectively), and the 50\% mask ratio stays within this noise band (40\% is the established block-masking optimum~\cite{bao2021beit,peng2022unified}). The mechanism-isolation deltas reported in the main paper compare configurations sharing the same pretraining randomness and are therefore not single-run artifacts.

\begin{table}[h]
\centering
\caption{\textbf{Seed and mask-ratio robustness} for the 2-expert Token+CLS model. Seeds 1--2 and mask=0.5 vs.\ the seed-42, mask-0.4 baseline. k-NN at $k\!=\!20$.}
\label{tab:robust}
\begin{tabular}{lccccc}
\toprule
 & Seed 1 & Seed 2 & Std & Mask=0.5 & Baseline \\
\midrule
k-NN & 75.40 & 75.23 & 0.078 & 75.34 & 75.39 \\
LP   & 79.54 & 79.63 & 0.037 & 79.23 & 79.60 \\
EP   & 80.21 & 80.28 & 0.039 & 79.80 & 80.19 \\
\bottomrule
\end{tabular}
\end{table}

\section{Token+CLS+Pixel Three-Objective Extension}
\label{sec:suppl_threeloss}

The main paper studies two-objective combinations (Token+CLS, Token+Pixel). We additionally trained the three-loss variant (2 experts, sparse, $\lambda\!=\!0.5$, $w_\text{pixel}\!=\!0.1$, otherwise matching the practical Token+CLS configuration). It reaches 75.15\% k-NN top-1, comparable to the matched non-MoE three-loss baseline (75.16\%). The third loss introduces gradient-magnitude conflicts that destabilize MoE specialization at default settings: $w_\text{pixel}\!=\!1.0$ underperforms by 0.67\%, and asymmetric per-expert entropy (the Token+Pixel-dense recipe) destabilizes dense-mode routing. At matched representation quality, per-patch routing in this configuration does not yet provide a measurable benefit over symmetric three-loss training; effective three-loss MoE requires per-loss expert-specialization analysis, which we leave to future work.

\end{document}